\documentclass[letterpaper, 10 pt, conference]{ieeeconf}  % Comment this line out if you need a4paper

\IEEEoverridecommandlockouts                              % This command is only needed if 
\usepackage{booktabs}
\usepackage{multirow}
\usepackage{amsmath} 
\let\labelindent\relax
\usepackage{enumitem}
\usepackage{amssymb}
\usepackage{graphicx} 
\usepackage{wrapfig}
\usepackage{float}      % for [H]
\usepackage{tcolorbox}
\usepackage[table]{xcolor}
\usepackage{algpseudocode}
\usepackage[ruled,vlined,linesnumbered]{algorithm2e}
\usepackage[font=small,labelfont=bf]{caption}
\usepackage[table]{xcolor}
\usepackage{url} 
\usepackage{hyperref}
\usepackage{cite}   
\usepackage{balance}

\title{\LARGE \bf
Smoothness as a Constraint for Stable Humanoid Locomotion
}

\author{ Utsav Panchal$^{1}$, Denis Kleyko$^{1}$, Unal Artan$^{1}$ and Amy Loutfi$^{1}$\\  
\thanks{$^{1}$AI, Robotics and Cybersecurity Center (ARC) and Department of Computer Science, Örebro University, Sweden.
        {\tt\small utsav.panchal@oru.se}}%
        \thanks{Project website: \url{https://panchalutsav.github.io/decap/}}%
}

\begin{document}
\bstctlcite{IEEEexample:BSTcontrol}

\maketitle
\thispagestyle{empty}
\pagestyle{empty}

\begin{abstract}
    Embodied AI systems, particularly humanoid robots deployed in real world scenarios require whole-body control policies that are both task-responsive and physically smooth. However, smoothness is not uniform across the body: lower body must remain sufficiently reactive, while the upper body must be tightly regulated to preserve stability. Existing reinforcement learning approaches typically impose smoothness through auxiliary terms in the reward function, which compete with task objectives, treating the body as uniform and provide no direct control over the physical quantities responsible for smooth behavior. We introduce DeCap (Decoupled Constraint-aware policy), a constrained reinforcement learning algorithm that decouples whole-body smoothness into separate upper- and lower-body constraint groups, each formulates smoothness as explicit constraints on physical motion limits. To improve constraint satisfaction near feasibility boundaries, DeCap incorporates a bounded barrier penalty that activates proactively as limits are approached and remains bounded at the constraint limit. On real-world humanoid whole-body control task, DeCap reduces upper-body action rate by 2.50$\times$ and acceleration by 2.18$\times$ relative to reward-based smoothness policies, while also improving lower-body smoothness and reducing  transient motion. We demonstrate that a fixed set of smoothness constraints transfers across diverse terrains, alleviating the need of extensive reward tuning. 

    \end{abstract}

\section{Introduction}

As humanoid robots leave controlled labs to more challenging real-world environments such as mines, warehouses, hospitals, and disaster zones~\cite{noreils2024humanoidrobotswork},  their motion is required to be smooth and physically safe. 
Abrupt actions, excessive joint accelerations, or high-frequency oscillations can induce instability or unsafe behavior~\cite{chen2024learningsmoothhumanoidlocomotion,lee2024gradientbasedregularizationactionsmoothness}, which all shorten the operational lifetime of hardware components~\cite{mysore2021regularizingactionpoliciessmooth}. 
To tackle this, current approaches focusing on stability~\cite{chen2024learningsmoothhumanoidlocomotion, li2025holdbeerlearninggentle} do not explicitly encode physical quantities that govern motion smoothness, and as a result they often struggle to produce smooth motion once deployed in the real world. Therefore, for humanoid control, motion smoothness should be treated as a safety- and hardware-critical requirement.

Real-world deployments further impose smoothness requirements that are not uniform across the body; rather, they are \emph{asymmetric} across body regions~\cite{li2025holdbeerlearninggentle,zhang2025falconlearningforceadaptivehumanoid}. The lower body must remain sufficiently reactive to absorb shocks from uneven terrain, whereas the upper body must remain stable to support: egocentric perception or payload carrying~\cite{lu2025mobile}.

\begin{figure}[!t]
    \centering
    \includegraphics[width=1.0\linewidth]{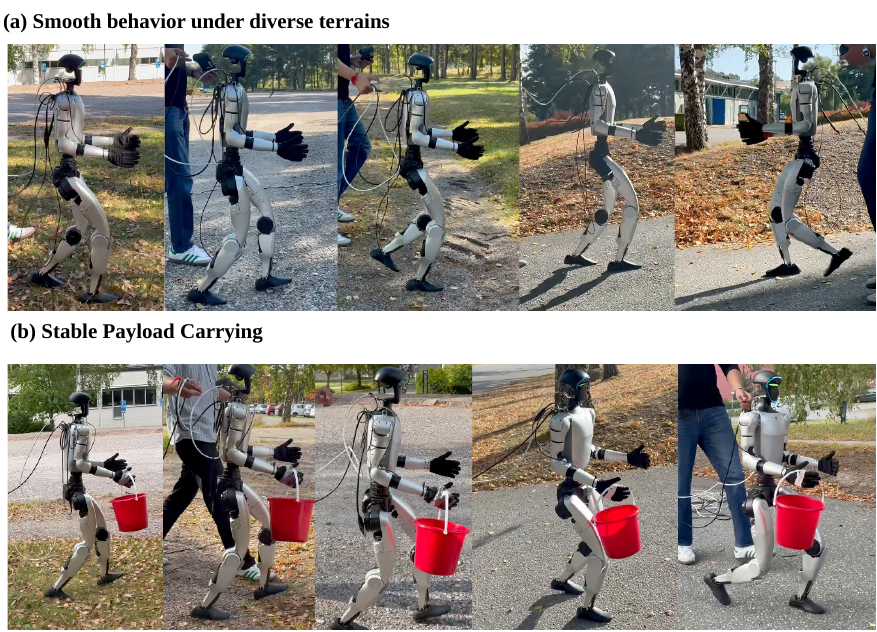}
    \caption{Our approach formulates whole-body smoothness as explicit physical constraints, with decoupled limits $\mathcal{E}_{\mathrm{upper}}$ and $\mathcal{E}_{\mathrm{lower}}$ for upper- and lower-body joint groups. The resulting policy is deployed zero-shot on the hardware, producing smooth behaviors  across diverse indoor and outdoor terrains: grass, gravel, slopes, and inverse slopes as well as during terrain transitions and payload carrying.}
    \label{fig:decap-terrains}
\end{figure}

Despite this need, previous approaches for humanoid whole-body control encode smoothness as an auxiliary reward term alongside task objectives such as velocity tracking or motion imitation~\cite{mysore2021regularizingactionpoliciessmooth,li2025holdbeerlearninggentle,rudin2022learningwalkminutesusing,liao2025beyondmimicmotiontrackingversatile,he2025asapaligningsimulationrealworld}. The policy is rewarded for being smooth, but it is not prevented from producing actions that exceed desirable velocity, acceleration, or jerk limits. This design fundamentally couples task performance and motion smoothness through hand-tuned weights.

This coupling could lead to two failure modes. First, the optimization is free to trade smoothness for primary task performance~\cite{chen2024learningsmoothhumanoidlocomotion} where the learned policy may emit high-acceleration actions that increase short-term reward but degrade safety, stability, and hardware longevity~\cite{lu2025mobile}. Second, reward weights that yield smooth whole-body control in one environment often fail to transfer to another~\cite{pmlr-v305-zhang25j, kim2024rewardsconstraintsapplicationslegged} because the underlying physical quantities such as joint acceleration, are not explicitly bounded~\cite{shin2025spectralnormalizationlipschitzconstrainedpolicies}.

These limitations raise two central challenges: \textit{i}) how to enforce smooth and physically safe motion without relying on reward engineering, and \textit{ii}) how to avoid environment-specific smoothness tuning while preserving the adaptability needed for RL-based humanoid whole-body control.

To address these challenges, we build on constrained RL (CRL), which offers a principled alternative to reward engineering~\cite{altman1999constrained}. Rather than including smoothness into the reward function, our approach formulates whole-body control as reward maximization subject to explicit smoothness constraints within the Constrained Markov Decision Process (CMDP) framework~\cite{altman1999constrained,achiam2017constrainedpolicyoptimization}. This decouples task performance from smoothness requirements: task rewards drive performance, while constraints define admissible motion. Our approach formulates motion smoothness as physical constraints, enabling different limits for upper- and lower-body dynamics and aims to reduce reward engineering. Experiments demonstrates that our approach can be a suitable alternative to reward-based~\cite{ben2025homie} or penalization based~\cite{chen2024learningsmoothhumanoidlocomotion} methods, offering direct control over physical limits without requiring smoothness to be encoded through manually tuned reward terms.

Our contributions are as follows:
\begin{itemize}
    
    % \item We introduce \textbf{DeCap} (\textbf{De}coupled \textbf{C}onstraint-\textbf{A}ware \textbf{P}olicy), a CRL algorithm that
    % formulates humanoid whole-body smoothness as explicit per-step constraints on
    % interpretable physical limits, rather than as competing terms in the reward function.
    \item We propose \texttt{DeCap} (\texttt{De}coupled
    \texttt{C}onstraint-\texttt{a}ware \texttt{P}olicy), which is built upon CRL with stochastic terminations~\cite{chanesane2024catconstraintsterminationslegged} and is
    formulated to enforce whole-body \emph{smoothness} on humanoid robots as constraints on
    interpretable physical quantities.

    \item We introduce decoupled body-group constraints by separating upper- and
    lower-body limits that enable reactive lower-body control while
    preserving upper-body stability.

    \item We formulate a bounded barrier penalty that activates proactively at
    a safety threshold, that remains bounded and continous, avoiding the divergence of classical log barriers.

    \item We perform a sim-to-real validation on a real-world humanoid across
    diverse terrains and an unseen payload (Fig.~\ref{fig:decap-terrains}),
    demonstrating stable locomotion and effective transfer.
     
\end{itemize}

\section{Related work}

\subsection{Stable Humanoid Whole-body Control} 

Humanoid whole-body control remains a fundamental challenge in robotics due to the high-dimensional and unstable dynamics of bipedal systems. Recent RL advances have significantly expanded the capabilities of legged robots by learning  policies for complex whole-body coordination directly through interaction with the environment~\cite{rudin2022learningwalkminutesusing}. Prior works have demonstrated robust whole-body behaviors including walking~\cite{rudin2022learningwalkminutesusing}, running~\cite{olkin2025chasingstabilityhumanoidrunning}, recovery from disturbances~\cite{margolis2025softmimiclearningcompliantwholebody}, terrain adaptation~\cite{kumar2021rmarapidmotoradaptation}, and agile maneuvers~\cite{he2025asapaligningsimulationrealworld, zhang2025hub}, while techniques such as domain randomization~\cite{rudin2022learningwalkminutesusing, campanaro2023learningdeployingrobustlocomotion} improve sim-to-real transfer by aligning dynamics in simulated and real-world environments.

Despite recent progress, achieving \emph{stable} humanoid whole-body control remains 
challenging, as RL policies optimized for task performance often result in 
oscillatory or jerky motions~\cite{chen2024learningsmoothhumanoidlocomotion, 
lee2024gradientbasedregularizationactionsmoothness}. Moreover, while recent 
work has extended humanoid locomotion to challenging outdoor 
terrain~\cite{radosavovic2024learning, 
Singh_2024}, these efforts target traversal 
robustness, and motion stability is typically evaluated only in controlled 
laboratory settings; whether learned policies remain \emph{smooth} on 
rough, or sloped surfaces is largely unexamined. Prior 
approaches encode smoothness through auxiliary reward 
terms~\cite{li2025holdbeerlearninggentle, 
huang2026steadytraylearningobjectbalancing}, but the resulting smoothness is 
an emergent byproduct of a weighted trade-off against task objectives rather 
than a specified physical quantity: as terrain difficulty increases, tracking 
terms dominate the fixed-weight penalties and smooth behavior silently 
degrades~\cite{pmlr-v305-zhang25j}. Because these weights carry no physical units, adapting them to new 
environments requires empirical retuning rather than principled 
adjustment~\cite{chen2024learningsmoothhumanoidlocomotion}. In contrast, we 
formulate smoothness as explicit constraints on physical quantities with 
interpretable bounds that remain valid across environments, and demonstrate 
that a single fixed constraint set produces stable whole-body control across five 
diverse real-world terrains.

\subsection{Constrained Reinforcement Learning} 

CRL is commonly formalized through the CMDP framework~\cite{altman1999constrained}, which augments the standard MDP with objective functions and explicit bounds, casting policy learning as reward maximization subject to objective constraints. A widely applied strategy is Lagrangian relaxation, which converts the constrained problem into an unconstrained problem via adaptive multipliers~\cite{tessler2018reward}. Recent approaches such as CPO~\cite{achiam2017constrainedpolicyoptimization} enforce constraints within each local update while IPO~\cite{liu2019ipointeriorpointpolicyoptimization} augments the objective with log barrier functions. These approaches provide principled constraint handling but are largely validated on simulated benchmarks without demonstrating sim-to-real transfer.

CRL has recently been adopted in legged robotics to encode safety constraints directly rather than through reward engineering. Lee~\emph{et~al.}~\cite{lee2024exploring} benchmark constrained policy optimization algorithms for quadrupedal locomotion and demonstrate that separating constraints from rewards streamlines training and improves sim-to-real transfer. The framework in~\cite{kim2024rewardsconstraintsapplicationslegged} reformulates most regularization terms as constraints handled by a modified IPO, while CaT~\cite{chanesane2024catconstraintsterminationslegged} enforces constraints through stochastic terminations, enabling agile quadrupedal skills. Naively transferring these formulations to humanoid robots is nontrivial because of the high-dimensional dynamics and different physical requirements which yield a substantially smaller feasible policy set, such that constraint configurations effective on quadrupeds often induce over-conservatism or convergence failure~\cite{huang2026eco}. Consequently, constrained formulations on humanoid hardware remains an open challenge. ECO~\cite{huang2026eco} enforces energy and gait-symmetry constraints via a Lagrangian method on a humanoid, while a related line of work targets motion smoothness through policy regularization, applying Lipschitz gradient penalties~\cite{chen2024learningsmoothhumanoidlocomotion} and spectral normalization~\cite{shin2025spectralnormalizationlipschitzconstrainedpolicies}; however, such regularizers bound policy sensitivity globally rather than constraining physical limits. While prior work has established constraint-based formulations as an effective 
tool for enforcing physical safety limits on legged robots, we show that the 
same formulation can be repurposed to enforce \emph{whole-body smoothness}: by 
expressing smoothness as per-step constraints, decoupled into separate 
upper-body and lower-body groups, our approach produces smooth motions that 
transfer to a Unitree G1 humanoid across diverse terrains.

\section{Method}
\subsection{Problem Formulation}

We formulate smooth humanoid whole-body control as an infinite horizon, discounted CMDP defined by $(\mathcal{S}, \mathcal{A}, r, \gamma, \mathcal{T}, {q_i})$, where $\mathcal{S}$ and $\mathcal{A}$ denote the state and action spaces, $r:\mathcal{S}\times\mathcal{A}\rightarrow\mathbb{R}$ is the task reward, $\gamma\in(0,1)$ is the discount factor, $\mathcal{T}: \mathcal{S} \times \mathcal{A} \rightarrow \mathcal{S}$ denotes the system dynamics, and $\{q_i(\cdot,\cdot):  \; i \in \mathcal{I}\}$ are constrained quantities.

The objective is to learn a policy $\pi:\mathcal{S}\rightarrow\mathcal{A}$ that maximizes the expected discounted return while satisfying constraint limits at every time step:

\begin{equation}
\begin{aligned}
\max_{\pi} \quad &
\mathbb{E}_{\tau \sim \pi, \mathcal{T}}
\left[
\sum_{t=0}^{\infty}
\gamma^t r(s_t, a_t)
\right] \quad 
\text{s.t.} \\ 
\quad &
% \mathbb{E}_{\tau \sim \pi,\mathcal{T}}
% \left[
% \sum_{t=0}^{\infty}
% \gamma^t c_i(s_t,a_t)
% \right]
% \leq \epsilon_i,
q_i(s_t, a_t) \leq \epsilon_i,
\;\; \forall i \in \mathcal{I},\; \forall t.
% \qquad \forall i \in \mathcal{I}.
\end{aligned}
\label{eq:rl_objective}
\end{equation}
\noindent
where $q_i(s,a)$ denotes the instantaneous value of the $i$-th constrained quantity and $\epsilon_i > 0$ its limit.
This formulation lets us specify desired behavioral properties (e.g., bounded joint accelerations or action smoothness) as explicit constraints rather than reward shaping.

\begin{figure}[t]
    \centering
    \includegraphics[width=1.0\linewidth]{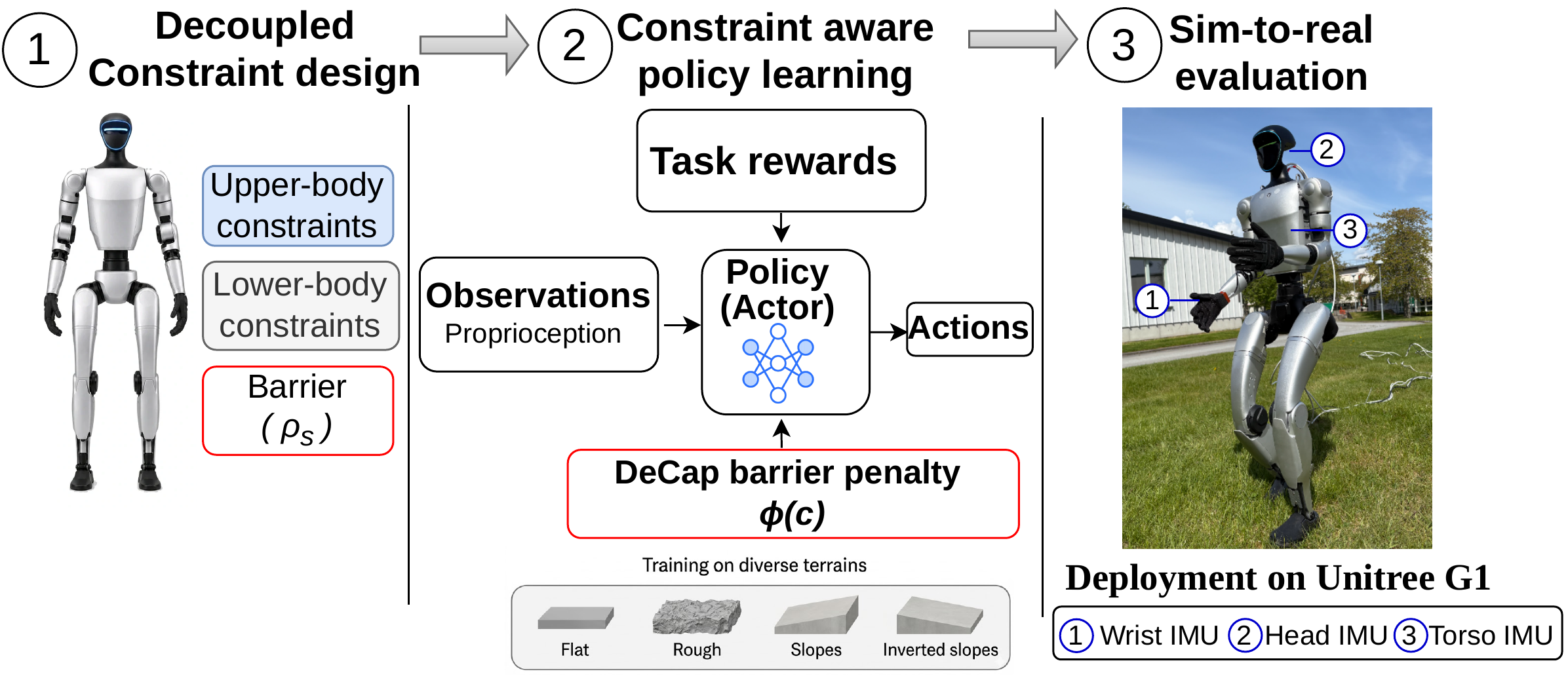}
    \caption{\textbf{DeCap overview.} An actor-critic policy is
    trained with privileged critic observations under smoothness
    constraints, with separate limits $\mathcal{E}_{\mathrm{upper}}$
    and $\mathcal{E}_{\mathrm{lower}}$ applied to upper- and lower-body
    joints. Training is randomized across terrains; the resulting
    policy is deployed and evaluated on the Unitree~G1 with three IMUs measuring stability.}
    \label{fig:saac-arch}
\end{figure}

\subsection{Constraints and Rewards}  
We define a set of constraints tailored to smooth humanoid whole-body control (listed in Table ~\ref{tab:constraints}). 
Rather than binary violation indicators used in previous works \cite{kim2024rewardsconstraintsapplicationslegged},  we define each cost as the constrained quantity itself, every time step yields a graded measure of violation magnitude and proximity to the limit.

For each constraint $i \in \mathcal{I}$, we specify a constraint limit $\epsilon_i$. To account for different physical tolerances of the humanoid's lower and upper bodies, we categorize these limits based on the joint group as shown in Fig.~\ref{fig:saac-arch}. \textbf{Upper body:} $\epsilon_i = \mathcal{E}_{\mathrm{upper}}$ for the shoulders, elbows, and wrists articulations and \textbf{Lower body:} $\epsilon_i = \mathcal{E}_{\mathrm{lower}}$ for the hips, knees, ankles, and waist articulations.

At each time step $t$, we compute two quantities $c_{i,t} = q_i(s_t, a_t) - \epsilon_i$ and $\hat{c}_{i,t} = q_i(s_t, a_t)/\epsilon_i$; 
$c_{i,t}$ indicates amount of constraint satisfied ($c_{i,t} < 0$) or violated ($c_{i,t} > 0$), whereas $\hat{c}_{i,t}$ expresses the proximity on a scale-free basis ($\hat{c}_{i,t} = 1$ at the boundary), making constraints with different
limits directly comparable.

\textbf{Selecting constraint limits.} A practical advantage of constraining 
physical quantities is that limits can be set directly from measured motion 
statistics. We profile the joint measurements of a nominal whole-body control policy, set 
$\mathcal{E}_{\mathrm{lower}}$ near the observed lower-body operating range 
to preserve reactivity, and tighten $\mathcal{E}_{\mathrm{upper}}$ towards the lower end of the observed upper-body range: for the locomotion and payload-carrying tasks, the nominal arm motion required for balance is small, and motion
beyond it primarily reflects policy oscillation rather than functional
movement.  

%--- Constraints -----------------------------------------------------
\begin{table}[t]
\caption{Constraint limits and termination probability bounds.
$\mathcal{E}_{\mathrm{upper}}$ and $\mathcal{E}_{\mathrm{lower}}$ denote the
limits $\epsilon_i$ applied to the upper- and lower-body joint groups,
and $p_i^{\max}$ is the maximum per-step termination probability of each
constraint (Eq.~\ref{eq:barrier}).}
\label{tab:constraints}
\centering
\footnotesize
\setlength{\tabcolsep}{5pt}
\begin{tabular}{@{}lcc@{}}
\toprule
& $\mathcal{E}_{\mathrm{upper}}$ & $\mathcal{E}_{\mathrm{lower}}$ \\
\textbf{Constrained quantity $q_i$} & {\scriptsize$(p_i^{\max}{=}0.50)$} & {\scriptsize$(p_i^{\max}{=}0.25)$} \\
\midrule
Action rate (rad/s)                & $5.0$ & $20.0$ \\
Joint acceleration (rad/s$^{2}$)   & $20.0$ & $600.0$ \\
Joint torque (Nm)                  & $4.0$  & $20.0$ \\
Joint velocity (rad/s)             & $1.5$  & $10.0$ \\
\bottomrule
\end{tabular}
\end{table}

In addition to the proposed constraints, we also employ a set of task-specific reward functions listed in Table~\ref{tab:rewards}. Unlike prior approaches ~\cite{li2025holdbeerlearninggentle, liao2025beyondmimicmotiontrackingversatile, huang2026steadytraylearningobjectbalancing}, \texttt{DeCap} does not include smoothness auxiliary terms in the reward function. 
% This removes the need for extensive smoothness reward engineering and hyperparameter tuning. 

%--- Rewards ---------------------------------------------------------
\begin{table}[t]
\caption{Key reward terms and weights used by DeCap algorithm.}
\label{tab:rewards}
\centering
\footnotesize
\setlength{\tabcolsep}{3pt}
\begin{tabular}{@{}lr@{\hskip 14pt}lr@{}}
\toprule
\textbf{Term} & \textbf{Weight} & \textbf{Term} & \textbf{Weight} \\
\midrule
Linear velocity (xy)  & $+1.5$  & Base height      & $-10.0$ \\
Angular velocity (z)  & $+1.0$  & Joint deviation  & $-1.0$  \\
Base linear velocity  & $-2.0$  & Foot slide       & $-0.2$  \\
Base angular velocity & $-0.05$ & Foot clearance   & $+1.0$  \\
\bottomrule
\end{tabular}
\end{table}

\subsection{Soft Barrier Penalty and Objective Formulation}
\label{subsec:barrier}

\begin{figure}[!t]
    % \vspace{14pt}
    \centering
    \includegraphics[width=0.9\linewidth]
    {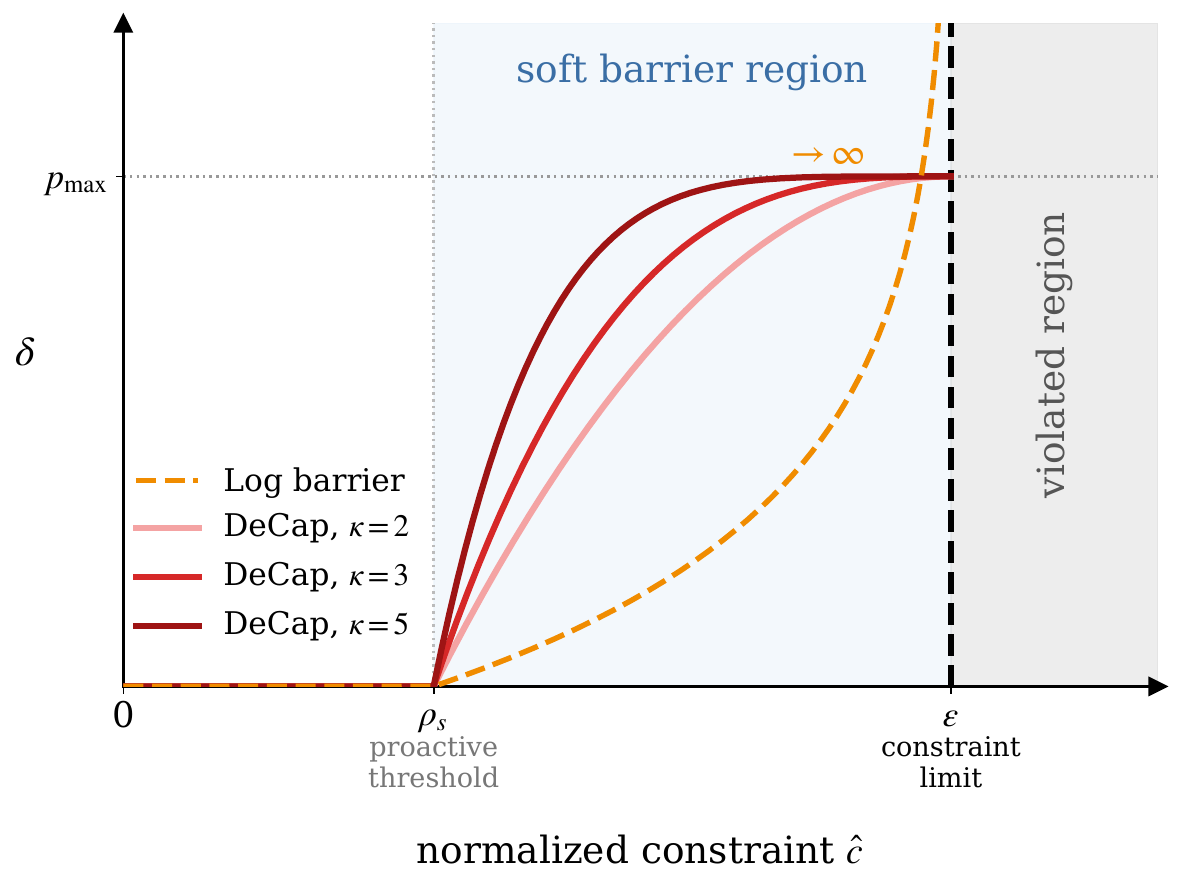}
    %\vspace{-6pt}
    \caption{\textbf{DeCap vs log barrier behavior at the boundary:} Our soft barrier penalty $\phi(\hat{c}_{i,t})$ increases smoothly as the normalized constraint $\hat{c}_{i,t}$ approaches the constraint limit.}
    % \vspace{5pt}
    \label{fig:saac-barrier}
\end{figure}

Constraint handling in legged locomotion largely follows two paradigms. \textit{Violation-triggered} methods, such as CaT~\cite{chanesane2024catconstraintsterminationslegged} and \textit{clamped penalties}~\cite{lee2024exploring}, produce a learning signal only after a constraint has been violated. For real-world deployment, the policy should respond preemptively as it approaches the constraint boundary~\cite{yang2025proactiveconstrainedpolicyoptimization, cheng2019endtoendsafereinforcementlearning}. Interior-point methods provide proactive signal through logarithmic barriers~\cite{liu2019ipointeriorpointpolicyoptimization}, but is unbounded as it approaches the constraint limit (see Fig. ~\ref{fig:saac-barrier}), requiring auxiliary mechanisms such as adaptive constraint thresholds~\cite{kim2024rewardsconstraintsapplicationslegged} to remain trainable from infeasible initializations. 

These limitations motivate a penalty that retains the proactive gradient while remaining bounded near the constraint limit. We design such a penalty on the normalized constraint
($\hat{c}_{i,t}$), whose boundary lies uniformly at
$\hat{c}_{i,t}=1$ for every constraint. We define a safety threshold
$\rho_s \in (0,1)$ that marks the onset of a margin preceding the boundary ($\hat{c}_{i,t}$) in which the
soft penalty activates (see Fig. ~\ref{fig:saac-barrier}): for the soft barrier region, i.e. $\rho_s \leq \hat{c}_{i,t} < 1$, we define the rescaled
coordinate:

\begin{equation}
    u_{i,t} = \frac{\hat{c}_{i,t} - \rho_s}{1 - \rho_s} \in [0, 1).
    \label{eq:rescaled}
\end{equation}

The classical log barrier $B(u_{i,t}) = -\log(1 - u_{i,t})$ provides the desired proactive signal but diverges as $u_{i,t} \to 1^{-}$, inducing numerical instability and excessively large gradients near the boundary. To retain its shape while removing the divergence, we compose it with the smooth, strictly increasing saturating map $\psi_\kappa(x) = 1 - e^{-\kappa x}$, which satisfies $\lim_{x \to \infty} \psi_\kappa(x) = 1$ and thus compresses the barrier's infinite range into $[0,1)$:
%\begin{equation}
\[
     \phi(\hat{c}_{i,t})\!=\!\psi_\kappa\big(B(u_{i,t})\big)\!=\!1\!-\!e^{-\kappa\left(-\log(1 - u_{i,t})\right)}\!=\!1\!-\!(1\!-\!u_{i,t})^{\kappa}
\]

Within the soft barrier region, the slope of $\phi(\hat{c}_{i,t})$ with respect to the
normalized constraint is
\begin{equation}
    \frac{\partial \phi}{\partial \hat{c}_{i,t}}
    = \frac{\kappa\,(1 - u_{i,t})^{\kappa - 1}}{1 - \rho_s},
    \label{eq:barrier_grad}
\end{equation}
\noindent
which, for $\kappa > 1$, $\phi(\hat{c})$ is concave and gradient decays smoothly to zero as $\hat{c}_{i,t} \to 1^{-}$ in contrast to the convex log barrier whose gradient diverges at the boundary. The value of $\kappa$ controls how sharply the penalty grows near the boundary (illustrated in Fig.~\ref{fig:saac-barrier}).  We set $\rho_s = 0.7$, so the policy encounters a soft repulsion once a constraint reaches $70\%$ of its limit $\epsilon_i$ (see ablations in Table~\ref{tab:constraint_summary}).

Beyond the boundary, we grade the penalty by the violation magnitude $c_{i,t}$, normalized by $\bar{c}_i$, an exponential moving average of the maximum violation across environments\cite{chanesane2024catconstraintsterminationslegged}. We combine
both regions into a per-step termination probability $\delta_{i,t}$, which defines the
probability that the episode terminates at step $t$ due to constraint $i$, thereby
forfeiting the future rewards~\cite{chanesane2024catconstraintsterminationslegged}. The per-step termination probability is 

\begin{equation}
    \delta_{i,t}\!=\!
    \begin{cases}
        0 & \!\!\hat{c}_{i,t}\!<\!\rho_s \\
        \phi(\hat{c}_{i,t})\,p_i^{\max} & \!\!\rho_s \leq \hat{c}_{i,t}\!<\!1 \\
        p_i^{\max}\!+\!(1\!-\!p_i^{\max})\,\mathrm{clip}\!\left(\dfrac{c_{i,t}}{\bar{c}_i},\! 0,\!1\right), & \!\!\hat{c}_{i,t}\!\geq\!1
    \end{cases}
    \label{eq:barrier}
\end{equation}
\noindent
where $p_i^{\max} \in (0,1)$ is the maximum penalty assigned before violation (see Table~\ref{tab:constraints}).

The policy then maximizes the expected discounted return under survival, optimized with PPO through termination-adjusted returns~\cite{chanesane2024catconstraintsterminationslegged}
\begin{equation}
    \mathcal{J}(\theta)\!=\!\mathbb{E}_{\pi_\theta}\!\left[\sum_{t=0}^{\infty} \gamma^tr_t
    \prod_{k=0}^{t-1}\big(1 - \bar{\delta}_k\big)\,\right],
    \quad \bar{\delta}_k\!=\!\max_i \delta_{i,k}.
    \label{eq:objective}
\end{equation}

\section{Experiments}

% We evaluate the performance of \texttt{DeCap} in both simulated and real-world environments. The experiments demonstrate that \textit{i}) in simulation \texttt{DeCap} leads to smoother whole-body control than reward-based approaches and   \textit{ii}) \texttt{DeCap} policies transfer to a Unitree~G1 humanoid (23 DoFs platform)~\cite{unitreerobotics} while maintaining smooth whole-body control across diverse terrains.

\subsection{Experimental Setup}
We train \texttt{DeCap} in the Isaac Lab simulator~\cite{mittal2025isaaclab}, which uses PPO~\cite{schulman2017proximal} as the underlying policy optimizer. 
Table~\ref{tab:ppo} presents PPO hyperparameters that are fixed across all evaluations. 
Table~\ref{tab:obs} lists the observation spaces: the critic
additionally accesses the privileged base linear velocity.
We incorporate domain randomization during training to improve sim-to-real transfer~\cite{rudin2022learningwalkminutesusing}.

% --- PPO hyperparameters ---------------------------------------------
\begin{table}[t]
\caption{PPO hyperparameters and training setup.}
\label{tab:ppo}
\centering
\footnotesize
\setlength{\tabcolsep}{3pt}
\begin{tabular}{@{}lr@{\hskip 14pt}lr@{}}
\toprule
\textbf{Hyperparameter} & \textbf{Value} &
\textbf{Hyperparameter} & \textbf{Value} \\
\midrule
Hidden layers & $[512,256,128]$ & Discount factor $\gamma$ & $0.99$ \\
Activation & ELU & GAE $\lambda$ & $0.95$ \\
Optimizer & Adam & Value loss coefficient & $1.0$ \\
Entropy coefficient & $0.01$ & Max gradient norm & $1.0$ \\
\bottomrule
\end{tabular}
\end{table}

%--- Observations ---------------------------------------------
\begin{table}[t]
\caption{Actor and critic observation spaces. The actor receives a partial observation, while the critic additionally has access to the privileged observation$^{\dagger}$ which is unavailable on hardware.}
\centering
\small
\setlength{\tabcolsep}{10pt}
\renewcommand{\arraystretch}{1.1}
\begin{tabular}{lcc}
\toprule
\textbf{Observation} & \textbf{Actor} & \textbf{Critic} \\
\midrule
Base linear velocity$^{\dagger}$ & & \checkmark \\
Base angular velocity            & \checkmark & \checkmark \\
Projected gravity                & \checkmark & \checkmark \\
Velocity command                 & \checkmark & \checkmark \\
Joint positions                  & \checkmark & \checkmark \\
Joint velocities                 & \checkmark & \checkmark \\
Previous action                  & \checkmark & \checkmark \\
\bottomrule
\end{tabular}

\label{tab:obs}
\end{table}

%--- Domain randomization --------------------------------------------
% \begin{table}[t]
% \caption{Domain randomization parameters during training.}
% \label{tab:dr}
% \centering
% \footnotesize
% \setlength{\tabcolsep}{5pt}
% \begin{tabular}{@{}lc@{}}
% \toprule
% \textbf{Parameter} & \textbf{Range} \\
% \midrule
% Static friction  & $[0.3,\,1.0]$ \\
% Base mass              & $[-1.0,\,+3.0]$\,kg \\
% Initial base position & $[-0.5,\,0.5]$\,m \\
% Initial joint velocity           & $[-1.0,\,1.0]$\,rad/s \\
% External push velocity ($x$, $y$) & $[-0.5,\,0.5]$\,m/s \\
% \bottomrule
% \end{tabular}
% \end{table}

For real-world deployment, we use Unitree G1 humanoid~\cite{unitreerobotics} as shown in Fig.~\ref{fig:saac-arch}. To assess upper-body stability and motion smoothness, we record inertial measurements from three IMUs: a torso-mounted Livox Mid-360 IMU (1050 Hz) and a head-mounted IMU (200 Hz) and an external Xsens
MTi-30 IMU mounted on the right wrist (400~Hz).

\begin{table*}[!tb]
\caption{Simulation results: performance of polices in IsaacLab on a flat terrain. All the policies are trained with three random seeds. The results are averaged on five evaluation rollouts over 256 environments of 1000 steps. Mean values are shown with standard deviations (\textcolor{gray}{gray}).}
\label{tab:metrics_sim}
\centering
\footnotesize
\setlength{\tabcolsep}{4.5pt}
\newcommand{\std}[1]{{\scriptsize\textcolor{gray}{$\pm#1$}}}

\begin{tabular}{llcccc}
\toprule
\textbf{Group} & \textbf{Method}
& \textbf{Action rate} $\downarrow$
& \textbf{DoF acceleration} $\downarrow$
& \textbf{DoF velocity} $\downarrow$
& \textbf{Energy} $\downarrow$ \\
\midrule

\multirow{6}{*}{Upper body} 
& \multicolumn{5}{l}{\cellcolor{gray!15}\textit{1. Reward-based methods}} \\
 
 & \texttt{Whole-body RL} & 
 $4.57$\,\std{0.80} & 
 $19.36$\,\std{0.25} & 
 $0.54$\,\std{0.16} & 
 $1.73$\,\std{0.19} \\

 & \texttt{Smoothness rewards} & 
$1.72$\,\std{0.15} &
$7.32$\,\std{0.17} &
$0.30$\,\std{0.10} &
$0.15$\,\std{0.13} \\

& \multicolumn{5}{l}{\cellcolor{gray!15}\textit{2. Constraint-based methods}} \\
& \texttt{LCP} &
$1.76$\,\std{0.27} &
$8.42$\,\std{0.15} &
$0.26$\,\std{0.31} &
$0.35$\,\std{0.26} \\

& \texttt{CaT} &
$1.63$\,\std{0.66} &
$7.19$\,\std{0.96} &
$0.28$\,\std{0.31} &
$0.27$\,\std{0.04} \\

& \texttt{N-IPO}&
$1.57$\,\std{0.15} &
$7.87$\,\std{0.04} &
$0.28$\,\std{0.09} &
$0.15$\,\std{0.03} \\

& \multicolumn{5}{l}{\cellcolor{gray!15}\textit{3. DeCap (ours)}} \\
 & \texttt{DeCap} w/o barrier  & 
 $0.83$\,\std{0.33} & 
 $4.00$\,\std{0.38} & 
 $0.32$\,\std{0.08} & 
 $0.20$\,\std{0.09} \\

 & \texttt{DeCap} w. $\kappa=2$ & 
 \textbf{0.65}\,\std{0.26} &
 \textbf{3.80}\,\std{0.33} &
 \textbf{0.16}\,\std{0.05} &
 \textbf{0.12}\,\std{0.05} \\

\midrule
\multirow{6}{*}{Lower body}

& \multicolumn{5}{l}{\cellcolor{gray!15}\textit{1. Reward-based methods}} \\
 & \texttt{Whole-body RL} & 
 $11.99$\,\std{0.76} &
 $44.61$\,\std{0.89} &
 $1.51$\,\std{0.05} &
 $16.77$\,\std{0.86} \\

 & \texttt{Smoothness rewards} &
$5.92$\,\std{0.10} &
$16.82$\,\std{0.17} &
$0.76$\,\std{0.03} &
$\mathbf{6.87}$\,\std{0.16} \\

& \multicolumn{5}{l}{\cellcolor{gray!15}\textit{2. Constraint-based methods}} \\
& \texttt{LCP} &
$8.75$\,\std{0.15} &
$31.96$\,\std{0.94} &
$1.11$\,\std{0.40} &
$10.41$\,\std{0.63} \\

& \texttt{CaT} &
$6.70$\,\std{0.60} &
$25.80$\,\std{0.79} &
$1.26$\,\std{0.94} &
$8.68$\,\std{0.65} \\

& \texttt{N-IPO} &
$9.06$\,\std{0.55} &
$35.35$\,\std{0.30} &
$1.17$\,\std{0.60} &
$8.27$\,\std{0.13} \\

& \multicolumn{5}{l}{\cellcolor{gray!15}\textit{3. DeCap (ours)}} \\
 & \texttt{DeCap} w/o barrier & 
 $5.99$\,\std{0.98} &
 $17.13$\,\std{0.03} &
 $0.79$\,\std{0.19} &
 $7.05$\,\std{0.38} \\

 & \texttt{DeCap}  w. $\kappa=2$ & 
 \textbf{5.02}\,\std{0.11} &
 \textbf{14.13}\,\std{0.40} &
 \textbf{0.75}\,\std{0.10} &
 $6.90$\,\std{0.21} \\
\bottomrule
\end{tabular}

\end{table*}

\subsection{Baselines}
% We compare \texttt{DeCap} with the following baselines which uses different approaches to enforce smoothness: 
We group baselines by how smoothness is enforced: \emph{reward-based} methods 
represent it through weighted penalty terms, whereas \emph{constraint-based} 
methods impose it through explicit or implicit limits on the policy. From each category, we select 
widely used methods that represent a distinct enforcement mechanism. 
% Together, these cover the 
% principal approaches to smooth whole-body control, allowing us to attribute 
% performance differences to the enforcement mechanism itself

\begin{itemize}[noitemsep, topsep=1pt, leftmargin=*]
    \item \texttt{Whole-body RL}: A reward-based approach, which uses first-order action rate penalty ($r_{\mathrm{smooth}} = ||a_t - a_{t-1}||$) adapted from~\cite{HumanoidVerse}. 
    \item \texttt{Smoothness rewards}: A reward-based approach from~\cite{ben2025homie}, where smoothness is enforced through a first and second-order action-rate penalties ($r_{\mathrm{smooth}} = ||a_t - a_{t-1}|| + ||a_t - 2a_{t-1} + a_{t-2}||$) in the reward function. 
    \item \texttt{LCP}: Lipschitz constrained policies adapted from~\cite{chen2024learningsmoothhumanoidlocomotion}.
    \item \texttt{CaT}: a constrained formulation adapted from~\cite{chanesane2024catconstraintsterminationslegged} with whole-body constraint limits. We use the lower-body limits $\mathcal{E}_{\mathrm{lower}}$ (Table~\ref{tab:constraints}) for the whole body. 
    \item \texttt{N-IPO}: Our reproduction of a constraint-based approach with log barrier formulation adapted from ~\cite{kim2024rewardsconstraintsapplicationslegged, liu2019ipointeriorpointpolicyoptimization}.
    \item \texttt{DeCap w/o barrier}: An ablation of our approach that keeps the decoupled constraint formulation but disables the soft barrier (i.e., $\phi \equiv 0$ for $\hat{c}_{i,t} < 1$).
\end{itemize}

\subsection{Metrics}
We evaluate performance using four smoothness metrics and report mean and standard deviation separately for upper- and lower-body joints: action rate (rad/s), DoF acceleration (rad/s$^2$), DoF velocity (rad/s),  and energy (N·rad/s).

\subsection{Results}
\subsubsection{Comparison with baselines} Table~\ref{tab:metrics_sim} compares 
\texttt{DeCap} against all baselines. Among the reward-based methods, strengthening the smoothness penalty from a
first-order action-rate term (\texttt{Whole-body RL}) to combined first- and
second-order terms (\texttt{Smoothness rewards}) reduces every metric across
both joint groups, confirming that reward engineering is effective when
carefully designed. The constraint-based 
baselines \texttt{LCP}, \texttt{CaT}, and \texttt{N-IPO} achieve comparable 
upper-body smoothness, but their lower-body results diverge: \texttt{LCP} and 
\texttt{N-IPO} retain high action rates ($8.75$ and $9.06$), while \texttt{CaT} 
improves to $6.70$. Since these methods enforce smoothness at the whole-body 
level, the policy remains free to trade smoothness in one body group for task 
returns in the other. \texttt{DeCap} instead assigns tighter limits to the upper body 
and looser limits to the lower body, which yields the lowest values across all upper-body
metrics. Lower-body improvements are smaller by 
design, yet \texttt{DeCap} still attains the best action rate and DoF acceleration, 
showing that decoupled constraints improve smoothness where it matters most 
without sacrificing lower-body responsiveness.

\subsubsection{Effect of barrier on joint behaviors} 
In Table~\ref{tab:metrics_sim}, we compare the two variants of \texttt{DeCap}: with and without the soft barrier formulation. The ablation remains within $1$--$4\%$ of the
\texttt{Smoothness rewards} baseline on the lower body while roughly halving
upper-body action rate ($0.83$ vs.\ $1.72$) and DoF acceleration ($4.00$
vs.\ $7.32$).  
% The 
% ablation matches the \texttt{Smoothness rewards} baseline on the lower body while achieving lower upper-body values. 
Activating the barrier at $\rho_s = 0.7$ of the limit 
instead penalizes the policy \emph{before} violations occur, pushing the action 
distribution into the interior of the feasible set. This proactive margin 
tightens every metric and the complete \texttt{DeCap} formulation thus achieves the lowest 
values across all metrics and joint groups.

\begin{figure}[t]
    \centering
    \includegraphics[width=1.0\linewidth]
    {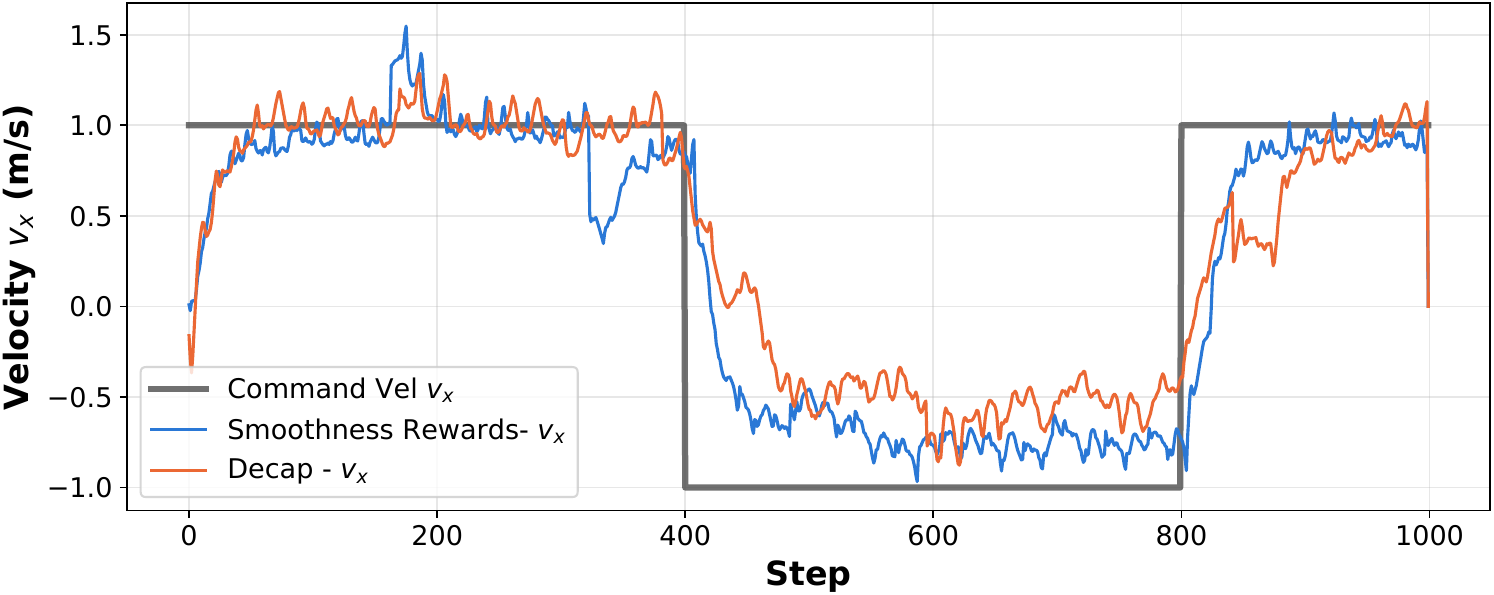}
    \caption{Velocity tracking of \texttt{DeCap} and \texttt{Smoothness rewards}.}
    \label{fig:vel_track}
\end{figure}

\subsubsection{Constraint satisfaction and task performance}  Fig.~\ref{fig:vel_track} shows the torso velocity tracking performance of
\texttt{DeCap} and the \texttt{Smoothness rewards} baseline under command switching and external
pushes applied every 4\,s during evaluation. \texttt{DeCap} closely tracks the commanded
forward velocity across command transitions while keeping relatively low violation rates (as shown in Table~\ref{tab:constraint_summary}) and consistently returns to the
reference after each push, demonstrating that constraint enforcement
does not come at the cost of tracking robustness or disturbance recovery.

\begin{figure}[!t]
    % \vspace{14pt}
    \centering
    \includegraphics[width=1.0\linewidth]
    {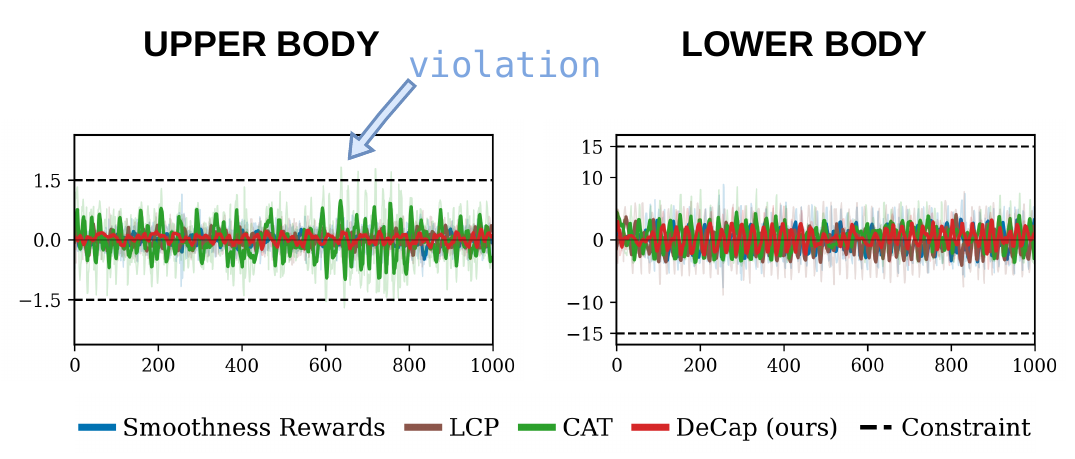}
    \caption{Per time-step traces for joint velocity for upper- and lower-body groups. Dashed lines indicates the constraint limit.}
    % \vspace{5pt}
    \label{fig:constraint_traces}
\end{figure}

% ------------------------------------------------------------
% Constraint Violation table
% ------------------------------------------------------------
\begin{table}[t]
\newcommand{\std}[1]{{\scriptsize\textcolor{gray}{$\pm#1$}}}
\caption{Constraint violation: average of time steps the constraints were violated by at least one joint, linear velocity tracking error and task return (\textit{xy}). Results are aggregated over five runs of 256 environments for 1000 steps.}
\label{tab:constraint_summary}
\centering
\setlength{\tabcolsep}{2.5pt}
\resizebox{\columnwidth}{!}{%
\begin{tabular}{lcccccc}
\toprule
Method & \multicolumn{2}{c}{Joint Acc. (\%) $\downarrow$} &
        \multicolumn{2}{c}{Action Rate (\%) $\downarrow$} & 
        \multirow{2}{*}{\shortstack{Vel. Track\\ (MAE) $\downarrow$}} & 
        \multirow{2}{*}{\shortstack{Task Return \\ (\textit{xy}) $\uparrow$}}\\
\cmidrule(lr){2-3}\cmidrule(lr){4-5}
& Upper & Lower & Upper & Lower & \\
\midrule

Whole-body RL
& $7.90$\std{0.10}
& $5.10$\std{0.09}
& $6.20$\std{0.12}
& $6.80$\std{0.10}
& $0.28$ 
& $39.02$\\

LCP
& $0.92$\std{0.04}
& $2.09$\std{0.00}
& $1.24$\std{0.03}
& $1.26$\std{0.08}
& $0.28$ 
& $30.73$\\

N-IPO
& $0.89$\std{0.12}
& $3.22$\std{0.17}
& $0.15$\std{0.07}
& $1.22$\std{0.19}
& $0.31$ 
& $32.02$ \\

CaT
& $0.31$\std{0.08}
& $0.18$\std{0.08}
& $1.43$\std{0.42}
& $0.25$\std{0.07}
& $0.29$ 
& $34.17$ \\

Smooth. rews.
& $0.15$\std{0.07}
& $0.16$\std{0.03}
& $0.18$\std{0.04}
& $0.31$\std{0.02}
& $0.30$ 
& $36.02$\\

DeCap ($\rho=0.7, \kappa=2$)
& ${0.10}$\std{0.05}
& $0.16$\std{0.02}
& ${0.12}$\std{0.01}
& ${0.13}$\std{0.01}
& $0.30$ 
& $35.00$  \\

DeCap ($\rho=0.7, \kappa=3$)
& $0.05$\std{0.02}
& $0.03$\std{0.02}
& $0.03$\std{0.01}
& $0.02$\std{0.01}
& $0.35$ 
& $34.95$  \\

DeCap ($\rho=0.7, \kappa=6$)
& $0.06$\std{0.02}
& $0.03$\std{0.01}
& $0.02$\std{0.01}
& $0.01$\std{0.01}
& $0.38$ 
& $34.50$  \\

DeCap ($\rho=0.9, \kappa=2$)
& $0.16$\std{0.04}
& $0.19$\std{0.03}
& $0.18$\std{0.05}
& $0.19$\std{0.03}
& $0.30$ 
& $37.20$  \\

DeCap ($\rho=0.5, \kappa=2$)
& $3.50$\std{1.05}
& $1.13$\std{0.03}
& $4.02$\std{1.09}
& $0.82$\std{0.07}
& $0.40$ 
& $27.45$  \\

\bottomrule
\end{tabular}%
}
\end{table}

Table~\ref{tab:constraint_summary} summarizes the average constraint violation
rates and the corresponding velocity tracking error. \texttt{Whole-body RL} has low tracking error but
violates $5$--$8\%$ of time steps. The \texttt{Smoothness rewards} baseline achieves comparable tracking performance with MAE of $0.30$, but violates action-rate limits by on $0.18\%$ (upper body) and $0.31\%$ (lower body). \texttt{LCP} retains a
competitive tracking performance, yet still exceeds
the lower body joint acceleration by $2.09\%$ of steps; since these
violations translate directly into jerky actuation, even these residual rates
are undesirable for achieving smooth behaviors in real world deployment. \texttt{CaT} demonstrates the limitation of enforcing constraints at the whole-body level, while it has relatively low violation on the lower-body, it still violates upper-body action-rate by $1.43\%$. This indicates that a single whole-body constraint formulation does not adequately capture the distinct constraint requirements of different body regions. \texttt{DeCap} decouples constraint groups and achieves lowest violation rates for joint acceleration while maintaining tracking performance comparable to the \texttt{Smoothness rewards} baseline, as also shown in Fig.~\ref{fig:vel_track}.

Table~\ref{tab:constraint_summary} also reports task returns for \textit{xy} velocity tracking; since each baseline encodes smoothness through different
reward terms while \texttt{DeCap} includes none, we compare only this shared
tracking component rather than total return. \texttt{Whole-body RL} attains the highest
return ($39.02$) by frequently violating smoothness limits, while
\texttt{LCP} and \texttt{N-IPO} achieve lower return with moderate
violation rates. \texttt{DeCap} attains a return of $35.00$, exceeding
\texttt{CaT} ($34.17$) and the other constraint-based baselines and
close to \texttt{Smoothness rewards} ($36.02$) baseline, indicating that comparable task performance can be achieved with substantially lower constraint violation rates.

Fig.~\ref{fig:constraint_traces} shows  per-time-step traces of joint velocity for both body groups,
with the constraint limits indicated by dashed lines. The upper-body joint
velocity plot makes the failure mode of baselines visible: it exhibits repeated
transient excursions beyond the constraint limit for the \texttt{CaT} approach, whereas the \texttt{DeCap} trace remains confined
within the bounds for the entire rollout.

\subsubsection{Effect of the tightness parameter \texorpdfstring{$\kappa$}{kappa} and barrier activation \texorpdfstring{$\rho$}{rho}} 
Table~\ref{tab:constraint_summary} reports the constraint violation rates and
linear velocity tracking error of \texttt{DeCap} trained with
different values of the barrier tightness parameter $\kappa$. Increasing $\kappa=2$ to $\kappa=3$ reduces constraint violations, but increases the velocity tracking error from $0.30$ to $0.35$ m/s -- highlighting the trade-off between constraint satisfaction and task performance. Further increasing $\kappa=6$ increases the tracking error to $0.38$ m/s and slightly lowers the task return to $34.95$. This degradation is consistent with the saturation of $\phi$ at large $\kappa$: beyond the immediate vicinity of
$\rho_s$ the penalty landscape becomes nearly flat as shown in Fig~\ref{fig:saac-barrier}, so the learning signal no
longer discriminates between states deep inside the soft barrier region and at the limit, which reduces the useful gradient of the barrier. We therefore keep $\kappa=2$ as a balanced setting that provides low violation rates while preserving tracking performance. 

We also ablate \texttt{DeCap} with different configurations of barrier activation ($\rho_s$) with fixed $\kappa=2$. Raising $\rho_s$ from $0.7$ to $0.9$ shrinks
the margin $[\rho_s, 1)$ for soft penalty, leaving less room to correct before a violation
occurs:  violation rates increase slightly (upper-body joint acceleration
$0.10\%\!\rightarrow\!0.16\%$; upper-body action rate
$0.12\%\!\rightarrow\!0.18\%$) while tracking error is unaffected. Setting $\rho_s$ to aggressively low value of $0.5$ leads the stochastic termination probability to ramp up immediately and the training collapses with high upper body acceleration violation rates. This indicates that early activation of soft penalty can destabilize the constraint-learning signal and produce worse compliance than the properly tuned barrier activation.

\begin{table}[!b]
\caption{Sim-to-real results from Unitree~G1: joint measurements are collected under identical walking trajectories and
averaged over three runs.}
\label{tab:metrics_real}
\centering
\footnotesize
\setlength{\tabcolsep}{3pt}
\newcommand{\std}[1]{{\scriptsize\textcolor{gray}{$\pm#1$}}}
\begin{tabular}{@{}lcccc@{}}
\toprule
\textbf{Method} & \textbf{Action\ rate $\downarrow$} & \textbf{DoF acc. $\downarrow$} & \textbf{DoF vel. $\downarrow$} & \textbf{Energy $\downarrow$} \\
\midrule
\rowcolor{gray!20}
\multicolumn{5}{@{}l}{\textbf{Upper-body}} \\

CaT  & 
$0.27$\,\std{0.15}          
& $3.77$\,\std{0.47}           
& $0.12$\,\std{0.18}          
& $0.12$\,\std{0.28} \\

Smooth. rews.  & 
$0.25$\,\std{0.12}          
& $3.82$\,\std{0.19}           
& $0.13$\,\std{0.02}          
& $0.14$\,\std{0.04} \\

DeCap (ours)     & 
$\mathbf{0.10}$\,\std{0.09} & $\mathbf{1.75}$\,\std{0.13}  & $\mathbf{0.11}$\,\std{0.05} & $\mathbf{0.10}$\,\std{0.03} \\

\midrule
\rowcolor{gray!20}
\multicolumn{5}{@{}l}{\textbf{Lower-body}} \\

CaT 
& $1.10$\,\std{0.94}          
& $11.20$\,\std{0.18}          
& $0.47$\,\std{0.60}          
& $4.09$\,\std{0.19} \\

Smooth. rews.  
& $1.04$\,\std{0.23}          
& $11.37$\,\std{0.42}          
& $0.48$\,\std{0.07}          
& $4.19$\,\std{0.40} \\

DeCap (ours)     
& $\mathbf{1.03}$\,\std{0.20} & $\mathbf{10.72}$\,\std{0.15} & $\mathbf{0.43}$\,\std{0.08} & $\mathbf{3.06}$\,\std{0.09} \\
\bottomrule
\end{tabular}
\end{table}

\subsection{Evaluation in real-world environments}
For real-world evaluation, we deploy one baseline from each category with the lowest overall constraint violation
rates as shown in Table~\ref{tab:constraint_summary}. 
\texttt{Smoothness rewards} among reward-based methods and \texttt{CaT} among
constraint-based methods. We deploy the trained policies on
the Unitree~G1 humanoid and report the results in
Table~\ref{tab:metrics_real}. The largest gains occur in the upper body, where both baselines behave similarly while \texttt{DeCap} reduces the mean action rate by $2.50$x
($0.25 \rightarrow 0.10$) and mean DoF acceleration by $2.18$x
($3.82 \rightarrow 1.75$) relative to \texttt{Smoothness rewards}. Upper-body mechanical
energy is reduced by 29\% ($0.14 \rightarrow 0.10$). Lower-body gains
are smaller by design, reflecting the intentionally looser constraints. Notably, \texttt{DeCap} also reduces
variability, suggesting that the learned policy
avoids sudden bursts of motion that excite hardware vibrations. This
trend matches our simulation results and shows that the constraint-based
formulation transfers effectively to real-world deployment.

\begin{table}[!t]
\caption{IMU Results from Unitree G1: root-mean-square (RMS) of the base angular velocity magnitude,
$\mathrm{RMS}=\sqrt{\tfrac{1}{T}\sum_{t=1}^{T}\lVert\boldsymbol{\omega}_t\rVert_2^2}$,
with $\boldsymbol{\omega}_t\in\mathbb{R}^3$ the gyroscope reading (rad/s) while locomotion,
measured by three independent IMUs on Unitree G1 as shown in Fig.~\ref{fig:saac-arch}.
Results are aggregated over the active-joystick window of each run
and three repeated trials per terrain (identical target commands and travel distance). \texttt{DeCap} achieves the lowest angular-velocity RMS across all terrains and IMU locations.}

\label{tab:imu_rms}
\centering
\footnotesize
\setlength{\tabcolsep}{2.5pt}
\renewcommand{\arraystretch}{1.0}
\newcommand{\std}[1]{{\scriptsize\textcolor{gray}{$\pm#1$}}}
\begin{tabular}{lccc}
\toprule
\textbf{Method} & \textbf{Head} $\downarrow$ & \textbf{Torso} $\downarrow$ & \textbf{Wrist} $\downarrow$ \\
\midrule

\rowcolor{gray!20}
\multicolumn{4}{l}{\textbf{ (a) Indoor Carpet}} \\
CaT  & 
$0.54$\std{0.15} & 
$0.50$\std{0.16} & 
$0.59$\std{0.18} \\

Smoothness rewards  & 
$0.40$\std{0.10} & 
$0.34$\std{0.15} & 
$0.51$\std{0.16} \\

DeCap (ours)        & 
$\mathbf{0.28}$\std{0.10} & 
$\mathbf{0.32}$\std{0.12} & 
$\mathbf{0.41}$\std{0.11} \\
\rowcolor{gray!20}

\multicolumn{4}{l}{\textbf{ (b) Outdoor Grass}} \\
CaT  
& $0.70$\std{0.15} & 
$0.66$\std{0.14} & 
$0.81$\std{0.11} \\

Smoothness rewards  & 
$0.60$\std{0.17} & 
$0.55$\std{0.15} & 
$0.63$\std{0.18} \\

DeCap (ours)  & 
$\mathbf{0.44}$\std{0.15} & 
$\mathbf{0.38}$\std{0.11} & 
$\mathbf{0.42}$\std{0.14} \\

\rowcolor{gray!20}

\multicolumn{4}{l}{\textbf{ (c) Outdoor Gravel}} \\
CaT  & 
$0.62$\std{0.31} & 
$0.59$\std{0.30} & 
$0.69$\std{0.35} \\

Smoothness rewards  & 
$0.61$\std{0.39} & 
$0.58$\std{0.37} & 
$0.68$\std{0.42} \\

DeCap (ours)  & 
$\mathbf{0.56}$\std{0.30} & 
$\mathbf{0.49}$\std{0.26} & 
$\mathbf{0.56}$\std{0.30} \\

\rowcolor{gray!20}

\multicolumn{4}{l}{\textbf{ (d) Slopes}} \\
CaT  & 
$0.57$\std{0.25} & 
$0.54$\std{0.24} & 
$0.69$\std{0.31} \\

Smoothness rewards & 
$0.53$\std{0.31} & 
$0.48$\std{0.29} & 
$0.59$\std{0.34} \\

DeCap (ours)  & 
$\mathbf{0.52}$\std{0.20} & 
$\mathbf{0.45}$\std{0.23} & 
$\mathbf{0.48}$\std{0.21} \\

\rowcolor{gray!20}

\multicolumn{4}{l}{\textbf{ (e) Inverse Slopes}} \\
CaT  & $0.58$\std{0.25} & 
$0.53$\std{0.25} & 
$0.63$\std{0.28} \\

Smoothness rewards  & 
$0.59$\std{0.31} & 
$0.51$\std{0.27} & 
$0.65$\std{0.32} \\

DeCap (ours) & 
$\mathbf{0.54}$\std{0.23} & 
$\mathbf{0.46}$\std{0.20} & 
$\mathbf{0.49}$\std{0.19} \\

\bottomrule
\end{tabular}

\end{table}

\begin{figure}[t]
    \centering
    \includegraphics[width=1.0\linewidth]
    {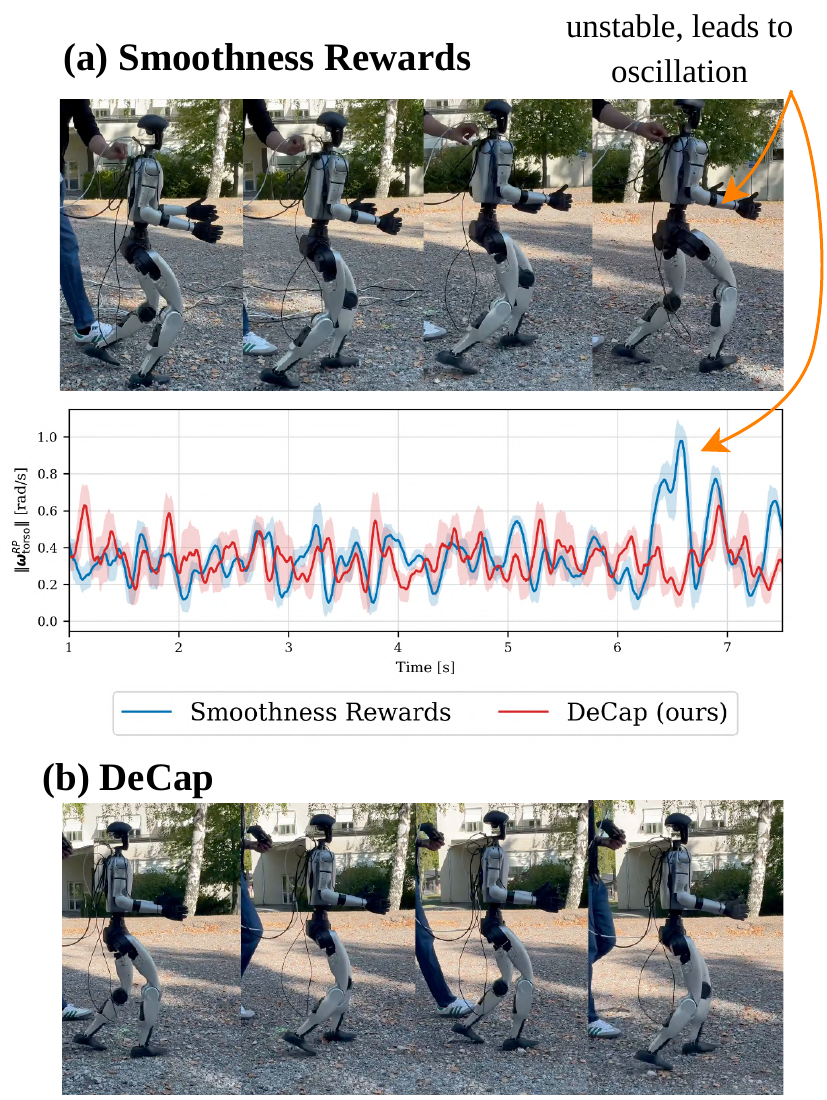}
    \caption{Torso IMU angular velocity over time on outdoor gravel.
The \texttt{Smoothness rewards} baseline exhibits intermittent
spikes, consistent with the higher RMS in
Table~\ref{tab:imu_rms}, whereas \texttt{DeCap} maintains a
smooth profile throughout the run.}
    % \vspace{5pt}
    \label{fig:diff-decap-sr}
\end{figure}

\subsubsection{Stability analysis across diverse terrains}

We evaluate the stability of the trained policies on the Unitree~G1
across five terrains: indoor carpet, outdoor grass, outdoor gravel,
slopes, and inverse slopes (see Fig.~\ref{fig:decap-terrains}).
Table~\ref{tab:imu_rms} reports the RMS of the base angular velocity ($|\boldsymbol{\omega}|$)
magnitude measured by three independent IMUs at the head, torso, and
wrist as shown in Fig.~\ref{fig:saac-arch}. \texttt{DeCap} reduces the angular-velocity RMS relative to the
baseline across all terrains. Reductions
relative to \texttt{Smoothness rewards} are
largest on indoor carpet and outdoor grass, and remain consistent at the wrist which is the most oscillation-prone location with reductions of $18$--$33\%$ across all
five terrains. \texttt{DeCap} also suppresses self-excited
oscillations of the policy. This is qualitatively visible in
Fig.~\ref{fig:diff-decap-sr}: on gravel terrain, the \texttt{Smoothness
rewards} policy exhibits intermittent spikes in the torso which destabilizes the motion. This is consistent with the observation that fixed-weight smoothness penalties do not explicitly bound the underlying physical quantities and can degrade under terrain-induced disturbances~\cite{kim2024rewardsconstraintsapplicationslegged}, whereas \texttt{DeCap} maintains a consistently smooth profile. The improved stability is also directly visible in
egocentric video from the head-mounted camera, provided on our project page.

\begin{figure}[!t]
    \centering
    \includegraphics[width=1.0\linewidth]
    {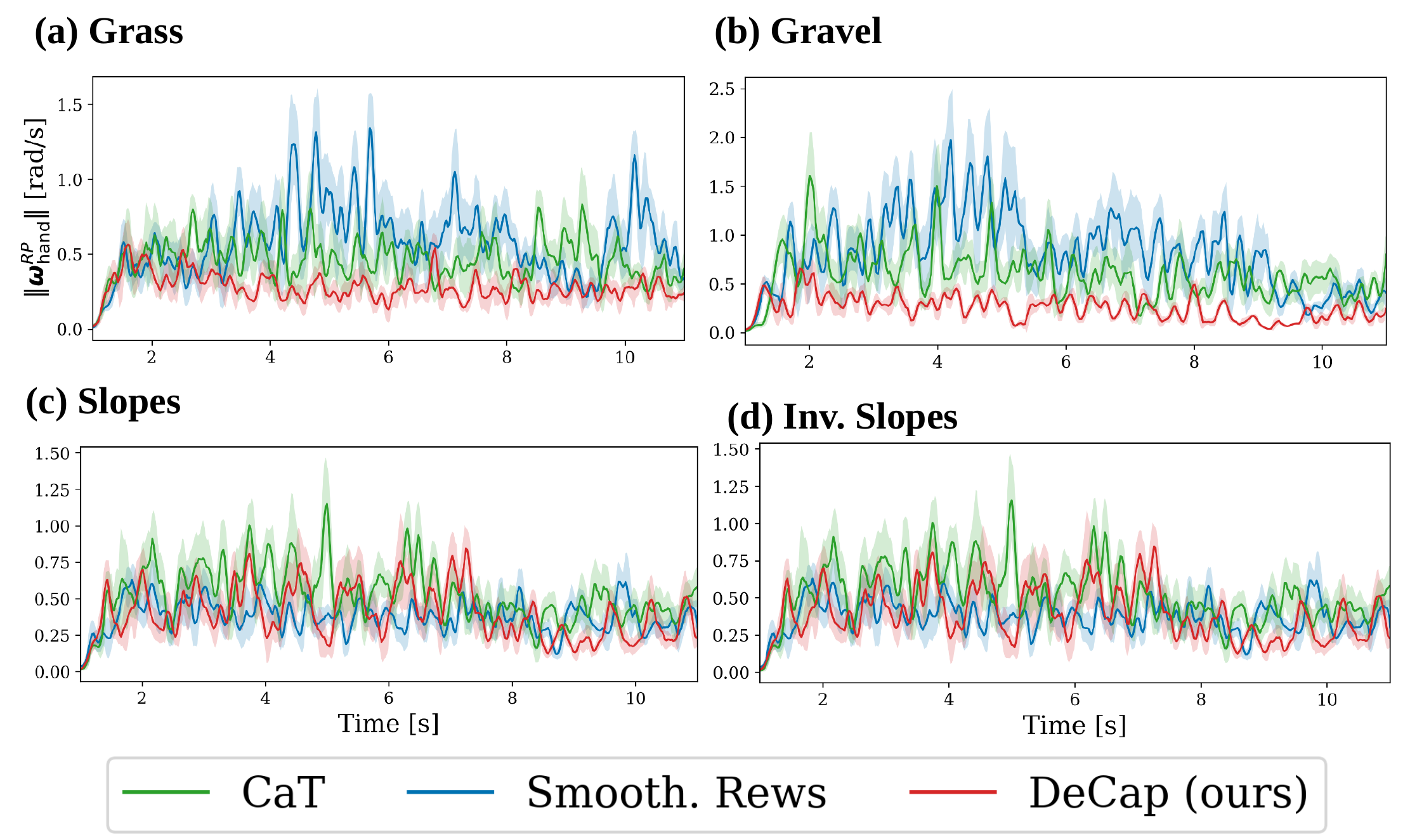}
    \caption{Right wrist IMU angular velocity over time during whole-body control
while carrying a 1.2\,kg payload. \texttt{DeCap} maintains an overall low,
stable profile throughout the run, whereas the baselines shows larger oscillations.}
    \label{fig:imu_payload}
\end{figure}

\subsubsection{Stable payload carrying}
We further evaluate robustness across all the terrains (Fig.~\ref{fig:decap-terrains}b) by attaching a
1.2\,kg payload to the right hand  and tracing the  angular velocity magnitude,
$\lVert\boldsymbol{\omega}_{\mathrm{hand}}^{\mathrm{RP}}\rVert
= \sqrt{\omega_x^2+\omega_y^2}$, from the wrist-mounted IMU. As shown in
Fig.~\ref{fig:imu_payload}, \texttt{DeCap} maintains a consistently low angular-velocity profile over extended runs, whereas the baselines exhibits larger and more frequent oscillations especially on gravel terrain. Notably, all \texttt{DeCap} runs were completed without operator intervention, while the baseline required the operator to intervene to prevent instability induced by the payload. Since both policies were trained with identical environment parameters, this gap indicates that the constraint-based formulation also generalizes to load conditions beyond those seen during training.

\section{Conclusion}

We presented \texttt{DeCap}, a constrained reinforcement learning algorithm for
smooth humanoid whole-body control with decoupled upper-body and lower-body
constraints. Unlike prior approaches, \texttt{DeCap} treats motion smoothness as
an explicit feasibility requirement rather than an auxiliary
reward-shaping term. Experiments in simulation and on the Unitree~G1
show consistent reductions in motion jitter and acceleration
variability over reward-based and constraint-based baselines, while
maintaining stable whole-body control across diverse terrains, demonstrating
that constraint-based smoothness is a practical alternative to reward
engineering for real-world humanoid control. \texttt{DeCap}'s main limitations
are the sensitivity of the soft barrier to the tightness parameter
$\kappa$ and the manual selection of constraint limits, which may
require re-tuning on hardware with different actuation
characteristics. Future work could integrate the constraint-based
formulation with imitation learning, or combine it with end-effector
and object-centric
rewards~\cite{li2025holdbeerlearninggentle, kim2024rewardsconstraintsapplicationslegged, 
huang2026steadytraylearningobjectbalancing} for loco-manipulation.

%                                   % the textheight of the last page by a suitable amount.
%                                   % This command does not take effect until the next page
%                                   % so it should come on the page before the last. Make
%                                   % sure that you do not shorten the textheight too much.

%%%%%%%%%%%%%%%%%%%%%%%%%%%%%%%%%%%%%%%%%%%%%%%%%%%%%%%%%%%%%%%%%%%%%%%%%%%%%%%%

%%%%%%%%%%%%%%%%%%%%%%%%%%%%%%%%%%%%%%%%%%%%%%%%%%%%%%%%%%%%%%%%%%%%%%%%%%%%%%%%



% \section*{ACKNOWLEDGMENT}

% The preferred spelling of the word ÒacknowledgmentÓ in America is without an ÒeÓ after the ÒgÓ. Avoid the stilted expression, ÒOne of us (R. B. G.) thanks . . .Ó  Instead, try ÒR. B. G. thanksÓ. Put sponsor acknowledgments in the unnumbered footnote on the first page.

\bibliographystyle{IEEEtran}
\balance
\bibliography{references}

\end{document}